\documentclass[conference]{IEEEtran}
\usepackage{graphicx}
\usepackage{citesort}
\usepackage{subcaption}

\usepackage{multirow}
\usepackage{url}
\usepackage{tabularx}

\begin{document}

\title{A Bootstrapped Model to Detect Abuse and Intent in 
White Supremacist Corpora}

\author{\IEEEauthorblockN{B. Simons}
\IEEEauthorblockA{School of Computing\\
Queen's University\\
Kingston, Canada\\
Email: b.simons@queensu.ca}
\and
\IEEEauthorblockN{D.B. Skillicorn}
\IEEEauthorblockA{School of Computing\\
	Queen's University\\
	Kingston, Canada\\
	Email: skill@cs.queensu.ca}
}

\maketitle

\begin{abstract}
Intelligence analysts face a difficult problem: distinguishing
extremist rhetoric from potential extremist violence.
Many are content to express abuse against some target group,
but only a few indicate a willingness to engage in violence.
We address this problem by building a predictive model for
intent, bootstrapping from a seed set of intent words, and
language templates expressing intent.
We design both an n-gram and attention-based deep learner 
for intent and use them as colearners to improve both the basis 
for prediction and the predictions themselves.
They converge to stable predictions in a few rounds. 
We merge predictions of intent with predictions of abusive
language to detect posts that indicate a desire for violent
action.
We validate the predictions by comparing them to crowd-sourced labelling.
The methodology can be applied to other linguistic properties
for which a plausible starting point can be defined.
\end{abstract}

\section{Introduction}

Intelligence analysts scan online data looking for the signals
of those who plan to carry out violent attacks.
Several models have been developed for what might be called
``violent thought" in islamist and white supremacist forums.
Detecting when violent thought transforms into violent
action is more problematic.
There are many examples of so-called ``armchair jihadists"
who post extensively but never do anything; and conversely
those who move very rapidly to violent action without
extensive discussion (for example, Farhad Khalil Mohammad Jabar,
who killed an Australian civilian police employee apparently
within hours of hearing a radical islamist talk).

Models that detect abusive language and hate speech are moderately
well developed. We expand their power by adding the capability to
detect intent.
Intent can be defined as ``the state of mind of one who aims
to bring about a particular consequence" \cite{oxford_reference_law_enforce}; 
when tied to abusive language, this acts as a signature for those 
who are most likely to carry out violent actions.
Intelligence analysts can use this abusive intent model to
focus attention on those whose posts make them of greatest concern.

Datasets labelled by intent do not exist (except perhaps in
classified environments).
We design a bootstrapped approach that starts from 
small, widely-agreed signals of intent from the literature;
and then bootstraps these into a pair of models, one
using n-grams and one using biLSTMs, to predict intent.
These are coupled with a deep-learning model of abusive
language to label posts in White Supremacist settings by
their abusive intent.
The intent model's predictions were compared to labels
generated by human volunteers, with more than 80\% agreement.

\section{Related work}

Linguistic understanding of intent begins with the work of
Leech  \cite{going_to_usage} who observed that \emph{will} and
\emph{going to} are the strongest signals of the
``future as outcome of present" \cite{verb_meaning}.
However, a distinction must be drawn between the future
as an outcome of present \emph{circumstances}, or present \emph{intentions},
and so the author's stance with respect to these
verbs is critical. In particular, the presence
of first-person pronouns distinguishes intent from observation.
Frame analysis examines the effect of communication on
audiences, and has been used by Sanfilippo 
\cite{violent_frames_in_action}
to assess the likelihood of violent action from online posts.
This approach has been partially automated \cite{vim_model}.

Intent has also been considered as a social process.
The relationship between an in-group and an out-group has
been leveraged to detect abusive language, and also to
understand and characterise threats \cite{intergroup_threat_theory}.
The cooccurrence of abusive language and intent has been
observed to be associated with violent actions
\cite{handbook,lone_wolf,lone_wolf_extension,rhetoric_translation_new_york}.

Detection of properties of interest in natural language either
uses bag-of-words approaches, or deep learning, mostly using
biLSTMs.
Intent detection has previously been studied in the context of
business interactions: what is this (potential) customer
trying/planning to do
\cite{microsoft_intent_classification,intent_in_queries,intent_in_discussion,intent_classification_twitter,purchase_intent,commercial_intent}.
Others have built models to detect intent in wider contexts such as
forums using patterns such as \emph{actionable verb object}
 \cite{open_intent_detection};
or regular expressions \emph{I \ldots want \ldots to \ldots}
 \cite{social_media_intent}.
The definition of intent used by these approaches is quite broad.

The detection of abusive language has been well-studied
\cite{hannah_thesis,storm_front_dataset,24k_tweet_dataset,24k_related_paper}
using both bag-of-words and deep-learning approaches, with
typical accuracies in the 90\% range.

A technique on which we build, called double bootstrapping,
was developed by Gao \emph{et al.} \cite{double_bootstrapping}.
For unlabelled datasets, it begins from a small dictionary of hate-speech 
terms, generates initial labels for the data,
feeds these into two models, which then co-train, each learning from the 
labels, making new predictions, and then passing the predictions to 
the other model.

\section{Methodology}

Three White Supremacists forums were used as datasets: Stormfront,
Iron March, and the manifesto of the New Zealand attacker in 2019.
As well, a Wikipedia dataset was used as a contrast set,
and an abusive language dataset used to train a predictor
for abuse.

Forum posts are noisy natural language, with misspellings
and typos, in-group language, and non-textual elements such as
emojis.
Each dataset was processed to remove quotations (detected from html tags),
user handles, html tags, emojis, and Unicode characters.
Hashtags were replaced by their content, divided into words if
camel case was used; characters repeated more than twice were replaced
by two occurrences.

Table~\ref{tb:processed_data_length} shows the effect of preprocessing.
The reduction in size for Stormfront reflects the large number
of quotations typically used, both from the other posts and
from media web sites.

\begin{table}[ht]
  \centering
  \caption{Dataset character lengths before and after processing}
  \label{tb:processed_data_length}

  \begin{tabular}{ | c | c | c | c | }
    \hline
    \textbf{Dataset} & \textbf{Unprocessed} & \textbf{Processed} & \textbf{Removed} \\ \hline
    Storm-Front (intent) & $252,968,165$ & $141,192,445$ & $44.1\%$ \\ \hline
    Wikipedia (intent) & $63,228,684$ & $6,0372,420$ & $4.5\%$ \\ \hline
    Abuse ensemble & $91,742,800$ & $87,291,993$ & $4.8\%$ \\ \hline
    Iron March & $9,668,405$ & $7,827,782$ & $19.0\%$ \\ \hline
    Manifesto & $98,642$ & $96,782$ & $1.9\%$ \\ \hline
  \end{tabular}
\end{table}

Documents were split into \emph{segments}, broken at sentence boundaries
or semicolons. Segments are the primary units for which predictions of
intent and abuse will be made.

A segment by word n-gram frequency matrix was created, using 
$n = 3$ to 6.
Examination of the cumulative frequency indicated that almost
nothing was lost by taking only the 500,000 most common n-grams.

Embeddings for the data from Stormfront were created
using FastText $0.9.2$ \cite{fast_text}
to produce $200$-dimensional vectors,
using skipgram with default training parameters.
The training process took roughly 5 days.

Deriving an embedding from the Stormfront dataset, rather
than using the generic embedding, shows the importance of
in-group language patterns.
Table~\ref{tb:ft_embeddings} shows the neighbours of ``liberal"
in the generic FastText embedding, while
Table~\ref{tb:sf_embeddings} shows the neighbours in
the embedding derived from Stormfront.
The default embedding shows a conventional political view
of ``liberal" while the customised embedding shows a much
more doctrinaire view.

\begin{table}[ht]
  \centering
  \caption{25 words closest to ``liberal" in default FastText embeddings \cite{fast_text_model}}
  \label{tb:ft_embeddings}

  \begin{tabular}{ | c | c | }
    \hline
    \textbf{Word} & \textbf{Cosine distance} \\ \hline
    conservative & 0.8223 \\ \hline
    liberals & 0.799998 \\ \hline
    leftist & 0.792378 \\ \hline
    ultra-liberal & 0.768488 \\ \hline
    left-liberal & 0.763335 \\ \hline
    hyper-liberal & 0.762267 \\ \hline
    right-wing & 0.754539 \\ \hline
    leftwing & 0.75325 \\ \hline
    left-wing & 0.752645 \\ \hline
    non-liberal & 0.751433 \\ \hline
    conservatives & 0.751097 \\ \hline
    liberalist & 0.750335 \\ \hline
    liberalism & 0.750089 \\ \hline
    moderate-liberal & 0.748242 \\ \hline
    libertarian & 0.745923 \\ \hline
    left-leaning & 0.742489 \\ \hline
    liberal- & 0.738087 \\ \hline
    ultraliberal & 0.725314 \\ \hline
    rightwing & 0.721381 \\ \hline
    liberal-minded & 0.715108 \\ \hline
    centrist & 0.714065 \\ \hline
    pro-liberal & 0.713632 \\ \hline
    liberal-progressive & 0.712557 \\ \hline
    pseudo-liberal & 0.705708 \\ \hline
    super-liberal & 0.705055 \\ \hline
  \end{tabular}
\end{table}

\begin{table}
\centering
  \caption{25 words closest to ``liberal" in customised 
FastText embedding built from Stormfront}
  \label{tb:sf_embeddings}
\begin{tabular}{ | c | c | }
    \hline
    \textbf{Word} & \textbf{Cosine distance} \\ \hline
    gliberal & 0.82885 \\ \hline
    leftist & 0.814532 \\ \hline
    liberalist & 0.803048 \\ \hline
    lliberal & 0.793655 \\ \hline
    multiculturalist & 0.792877 \\ \hline
    libtard & 0.791708 \\ \hline
    leaningliberal & 0.791068 \\ \hline
    ultraliberal & 0.784252 \\ \hline
    libtarded & 0.77373 \\ \hline
    lberal & 0.763863 \\ \hline
    lefty & 0.762339 \\ \hline
    liberalistic & 0.760414 \\ \hline
    liberals & 0.758157 \\ \hline
    iliberal & 0.757605 \\ \hline
    socioliberal & 0.757185 \\ \hline
    conservativeliberal & 0.756382 \\ \hline
    milticulturalist & 0.741517 \\ \hline
    liberall & 0.739852 \\ \hline
    ultraliberals & 0.739685 \\ \hline
    ozliberal & 0.738213 \\ \hline
    liberalminded & 0.737438 \\ \hline
    egalitarian & 0.735231 \\ \hline
    muticulturalist & 0.734759 \\ \hline
    liiberal & 0.734605 \\ \hline
    justaliberal & 0.734268 \\ \hline
  \end{tabular}
\end{table}

\subsection{Intent label inference}

\begin{figure}[ht]
  \centering
  \includegraphics[width=.45\textwidth]{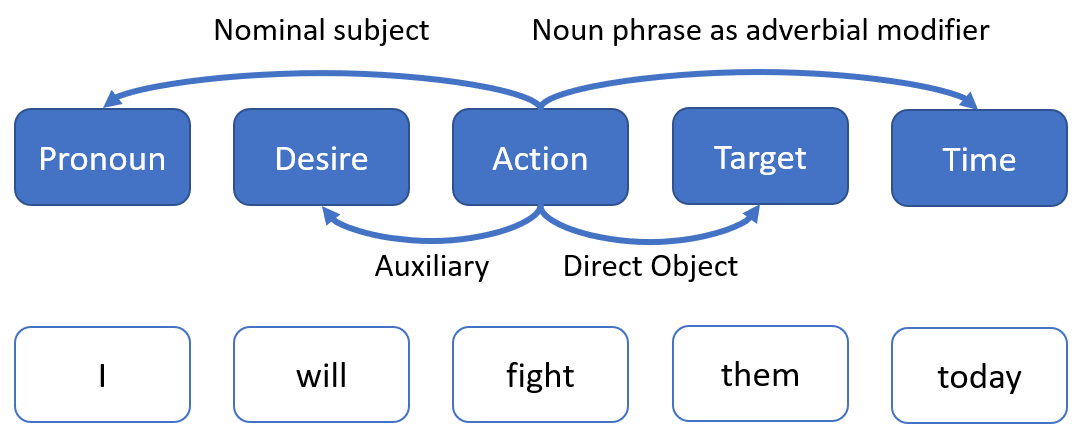}
  \caption{Short form of explicit intent with example}
  \label{fig:short_intent}
\end{figure}

\begin{figure}[ht]
  \centering
  \includegraphics[width=0.45\textwidth]{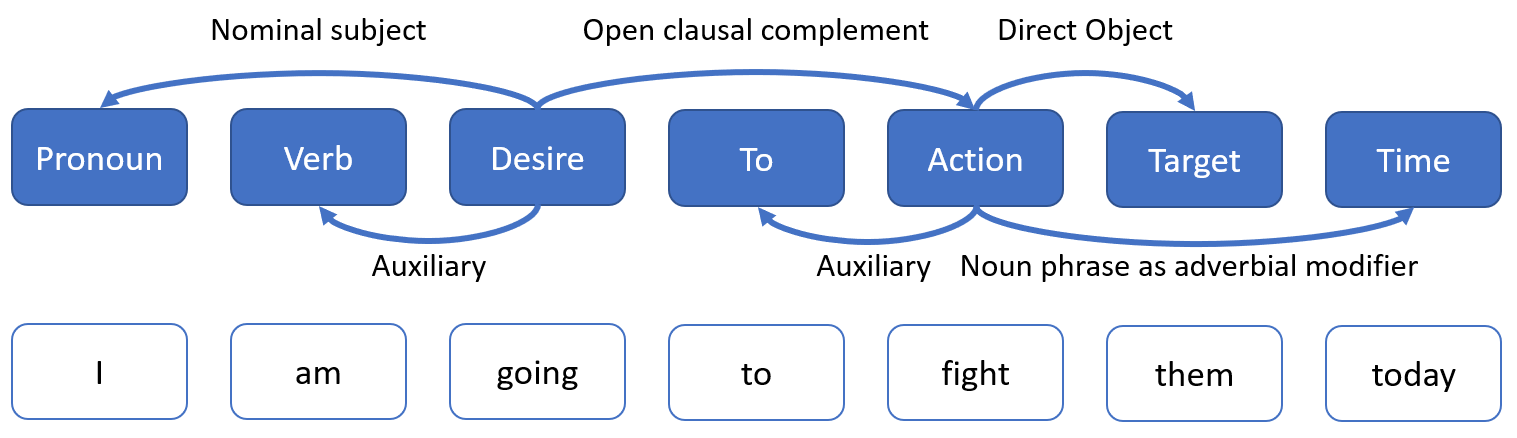}
  \caption{Long form of explicit intent with example}
  \label{fig:long_intent}
\end{figure}

Datasets labelled with intent are not publicly available, so we
develop a way of inferring intent labels by bootstrapping.
The first step is to develop a template for expressing
intent.
The basic structure of the template is: 
first-person pronoun, desire verb (``going to",
``will"), action verb (``fight") and optionally a target and
timing.
The template elements must occur in the appropriate order and
relationships.
Figures~\ref{fig:short_intent} and \ref{fig:long_intent} and 
Tables~\ref{tb:short_template} and \ref{tb:long_template}
provide examples.

\begin{table}
  \centering
  \caption{Components of the short form intent template} \label{tb:short_template}
  \begin{tabular}{ | c | c | c | }
    \hline
    \textbf{Role} & \textbf{Parent} & \textbf{Relationship to parent} \\ \hline
    Pronoun & Action verb & Nominal subject \\ \hline
    Desire verb & Action verb & Auxiliary \\ \hline
    Action verb & None & N/A \\ \hline
    Target (Optional) & Action verb & Direct object \\ \hline
    Timing (Optional) & Action verb & Noun phrase as adverbial modifier \\ \hline
  \end{tabular}
\end{table}

\begin{table}
  \centering
  \caption{Components of the long form intent template} \label{tb:long_template}

  \begin{tabular}{ | c | c | c | }
    \hline
    \textbf{Role} & \textbf{Parent} & \textbf{Relationship to parent} \\ \hline
    Pronoun & Action verb & Nominal subject \\ \hline
    Desire verb & None & N/A \\ \hline
    To & Action verb & Auxiliary \\ \hline
    Action verb & Desire verb & Open clausal complement \\ \hline
    Target (Optional) & Action verb & Direct object \\ \hline
    Timing (Optional) & Action verb & Noun phrase as adverbial modifier \\ \hline
  \end{tabular}
\end{table}

Segments from the dataset are parsed using the spaCy statistical parser
\cite{spacy} which is able to handle the variability of online documents,
including misspellings. It produces POS tags and dependencies.

Initial labels for each segment are generated as follows: if it contains
a match for the template, it is labelled $+1$; if it contains a negation
question, or a second- or third-person pronoun, it is labelled $0$;
otherwise it is labelled $0.5$.
A set of segments from Wikipedia was included in the set
as well, all labelled with $0$.
Wikipedia's style guidelines disallow instances of 
first-person intent \cite{wikipedia_guidelines} so this enhances the
dataset with examples of (highly likely to be) non-intent segments.

The desire verbs in the template can be considered as varying from
weak ones (dreaming of -- ```I am dreaming of attacking them")
to strong (have to -- ``I have to attack them").
In the initial round, any verb that is used in the desire verb slot
is counted as a desire verb, although the obvious choices
-- ``going to", ``will" -- were by far the most common.

The initial label generation phase, applied to the segments
from Stormfront and Wikipedia labelled 1.4\% as intentful, 29.1\%
as non-intentful, and 69.5\% as undetermined. As expected,
all of the Wikipedia segments were labelled as non-intentful.

The set of desire verbs is now refined.
This requires both expansion so that misspellings and other
variants are captured; and restriction so that only strong
desire verbs remain. Fortunately, using the geometry of the
embedding space solves both problems.

The process is initialised from a seed set of ``obvious" desire verbs:
``want", ``need", ``going", ``have", ``about", ``planning", 
and ``will".
The embedding is treated as a vector space and verbs
derived hypercone prism based on the seven 
initial vectors are treated as a strong desire verbs.
To be included, the unnormalized cosine similarity of a
selected desire verb must be less than twice the distance to the 
mean of the seed set than at least one member of the set.
This produces 596 desire verbs, including verbs such as
``seek" and ``aiming". 95.3\% of the segments
initially labelled as intentful still qualify.

\subsection{Bootstrapping using two concurrent learners}

The model is based on two learners:
an n-gram based learner;
and a deep learner that uses
the FastText embeddings and 
a BiLSTM-attention architecture \cite{bidirectional,hierarchical_attention}.
These are used in parallel, and the results of each better
inform the signals of intent for the next round.

\subsubsection{n-gram learner}

Assume that we have a set of label values between 0 and 1 for the segments; 
(initially this is obtained from the template matching).
The n-gram predictor first assigns a score to each n-gram based
on the ratio of how often it occurs in segments whose label
is greater than 0.5 versus how often it occurs in segments
whose label is less than 0.5, 
both frequencies normalized by the number of such segments in which
it is present.
n-grams in the 99.9th percentile of ratio values in either direction
are considered predictive of intent or non-intent.

Segments are labelled as intent if they contain only intentful
n-grams; as non-intent if they contain only non-intentful n-grams,
but unlabelled if both or neither happen to be present.

\subsubsection{Deep learner}

In parallel, a deep learner using a biLSTM is applied to the
segments, knowing the FastText embeddings of each word.
The network architecture is shown in Figure~\ref{tb:intent_architecture}.
The biLSTM begins with a conventional learning phase in
which it trains based on a randomly selected set of
segments with the labels for each segment rounded to 0 or 1).
As it trains the loss contribution from each segment is downweighted 
by how far its current label is from either 1 or 0.

After this phase, the deep learner then predicts labels for
all of the segments, including those without labels.
Segments close to the extremes have their scores altered by 10\%
to make them even stronger.

\begin{table}[ht]
  \centering
  \begin{tabular}{| c | c | c | c |}
    \hline
    \textbf{Layer} & \textbf{Units} & \textbf{Input dimension} & \textbf{Output dimension} \\ \hline
    BiLSTM & $200$ & $200 \times 200$ & $200 \times 400$ \\ \hline
    Attention & $400$ & $200 \times 400$ & $400$ \\ \hline
    Dense & $50$ & $400$ & $50$ \\ \hline
    Dense & $1$ & $50$ & $1$ \\ \hline
  \end{tabular}
\caption{biLSTM architecture}
\label{tb:intent_architecture}
\end{table}

\subsubsection{Merging predictions for the next round}

Both learners are trying to increase the number of strong predictions,
that is predictions close to 1 or close to 0.
However, they are simultaneously learning the natural language
signals of intent.
To prevent too rapid convergence of labels, each model is limited
in how many labels it can present to the merge mechanism.

If the previous round of each model had predicted $n_p$ labels with
value 1 and $n_n$ labels with value 0, then at most these numbers
of further records can be suggested to the merge mechanism. In other
words, if 100 labels were 1 in the previous round, then at most 200 segments
can be presented as 1 to the merge mechanism in the current round.

This limitation on convergence could potentially slow down the
overall process when a model quickly discovers the required
signals, However, this does not seem to be the case for complex properties
such as intent.

The merge mechanism relabels each segment based on the label
from either of the models, except if they have contradictory senses
(that is, one predicts intent while the other predicts non-intent)
in which case the label remains unchanged.
If a segment is labelled either 0 or 1, that label is locked and will
not change in subsequent rounds.

The number of rounds can be altered for each particular dataset, but
5 or 6 seems to be adequate.

\section{Modelling abusive language}

There is intrinsic interest in modelling intent, but the
applications we envisage are those where intent per se
is not interesting (``I'm going to buy her a birthday present")
but where intent associated with abusive language signals
the potential for violent action.
We must therefore also identify segments that express abuse.

Abusive language detection is a well-studied problem and
labelled data is available \cite{hannah_thesis,hierarchical_attention}.
We use three relevant datasets:
a small set of documents from Stormfront; 
a corpus from a competition run by Impermium for detecting insults 
\cite{insults_dataset};
and a dataset from a competition by Conversation AI to identify 
multiple types of abusive language in 
Wikipedia comments \cite{kaggle_dataset}.
The documents from these datasets were randomly mixed to ensure no 
sequential structure, creating a dataset with $240,846$ samples 
(Table \ref{tb:abuse_dataset}).

\begin{table}[ht]
  \centering
  \caption{Abusive language dataset composition} \label{tb:abuse_dataset}

  \begin{tabular}{| c | c | c |}
    \hline
    \textbf{Dataset} & \textbf{Size} & \textbf{Abusive Fraction} \\ \hline
    Storm-Front & $10,703$ & $11.2$ \% \\ \hline
    Insults & $6,594$ & $26.4$ \% \\ \hline
    Wikipedia comments & $223,549$ & $9.9$ \% \\ \hline
    \hline
    Ensemble & $240,846$ & $10.4$ \% \\ \hline
  \end{tabular}
\end{table}

A biLSTM predictive model, using the word embeddings
derived from Stormfront and 
the same network as for intent (Table~\ref{tb:intent_architecture}).

An 85--15 train-test split (204,719 training segments, 36,127 test
segments) was used, with a maximum of 50 epochs, and early stopping
with patience 3.
The learning rate was $0.001$, beta one and two were $0.9$ and $0.999$, 
and epsilon was $1 \times 10^{-7}$.
This model's accuracy was 86.7\%.

\subsection{Combining intent and abuse scores}

Both models produce predictions in $[0, 1]$ and there are multiple
ways in which these could be combined.
For our application domain, where the primary goal is to identify
segments that are high in both intent and abuse, we multiply
the two predicted scores.
Even moderate scores in either dimension reduce the overall
score, focusing attention on segments of greatest practical interest.

\subsection{Scores for documents from scores for segments}

Combining scores for segments to produce a single score for
each document raises subtle problems.
An author may be abusive in one segment and segue into intent
in the adjacent one, or \emph{vice versa}.
A document may begin with a discussion of a perceived problem
(neither abusive or intentful) and only then begin to express
abuse or intent or both.

The approach we chose was to consider the abuse and intent
scores of each segment separately, take each set of three
adjacent (overlapping) segments in a document, 
take the maximum abuse score and the maximum intent score in
the set of three, and then form the product of these maximums.
The overall score for the document is the maximum of these
products over all of the three-segment windows.

\section{Performance and validation}

In each epoch of intent training, the sequence learner and the
deep learner are run in parallel, and then a consensus is derived.
Table~\ref{tb:positive_rates_end} shows some of the highest
scoring intentful sequences after 6 epochs of training.
In general, both the set of strongest intentful and non-intentful
sequences converge quickly.

\begin{table}[ht]
  \centering
  \caption{15 sequences with the highest intentful rate after epoch 6}
  \label{tb:positive_rates_end}

  \begin{tabular}{ | c | c | }
    \hline
    \textbf{Sequence} & \textbf{Intentful Rate} \\ \hline
    i want to know & 220.17 \\ \hline
    that we need to & 175.59 \\ \hline
    i must say & 164.11 \\ \hline
    and we need to & 163.44 \\ \hline
    i will give & 159.38 \\ \hline
    i must admit & 158.03 \\ \hline
    i ll go & 141.82 \\ \hline
    but i want & 135.07 \\ \hline
    i ll post & 125.62 \\ \hline
    must say that & 122.91 \\ \hline
    we don t need to & 122.91 \\ \hline
    i ll just & 118.86 \\ \hline
    i must say that & 117.51 \\ \hline
    i will make & 109.41 \\ \hline
    i ll keep & 108.06 \\ \hline
  \end{tabular}
\end{table}

\begin{figure}[ht]
    \centering
    \includegraphics[width=0.45\textwidth]{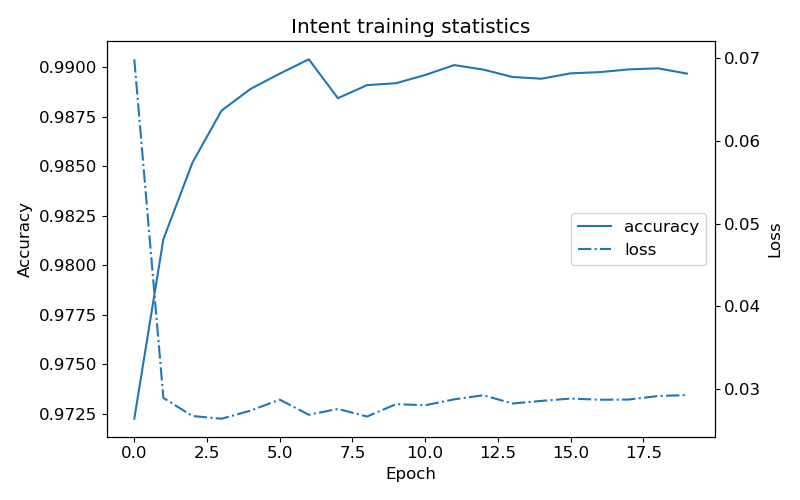}
    \caption{Accuracy and loss for the deep learner for intent}
    \label{fig:intent_training}
\end{figure}

Figure~\ref{fig:intent_training} shows the convergence of the deep
learner for intent. This figure should be interpreted with
care because the labels are changing from one epoch to the next,
but the convergence is clear.

\begin{figure}[ht]
    \centering
    \includegraphics[width=0.45\textwidth]{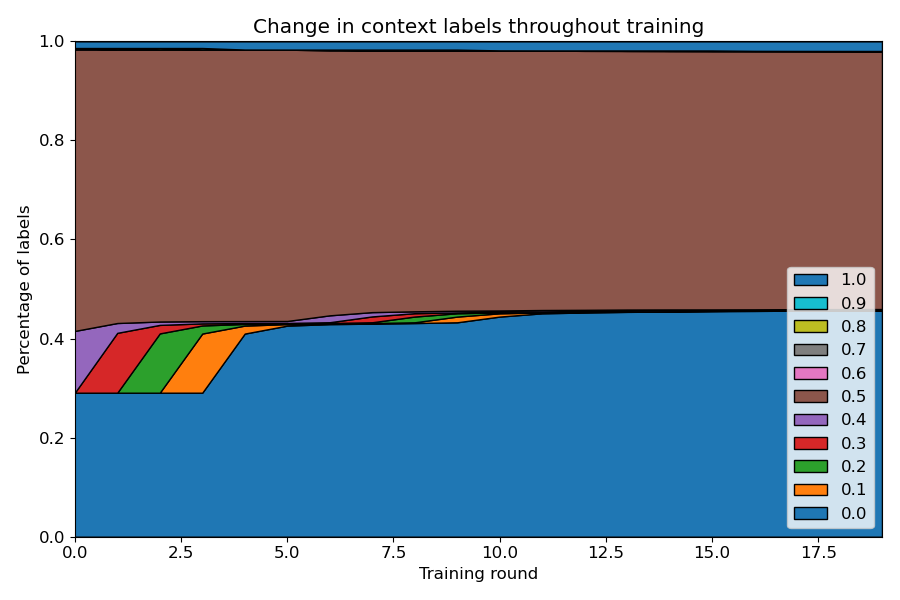}
    \caption{Visualization of the consensus labels after each training epoch}
    \label{fig:label_movement}
\end{figure}

The effect of each epoch on the label consensus is shown in
Figure~\ref{fig:label_movement}. The set of high-intent segments
do change, but not enough to be visible at the top of the figure; the
non-intent segments rapidly increase.

\begin{figure}[ht]
    \centering
    \includegraphics[width=0.45\textwidth]{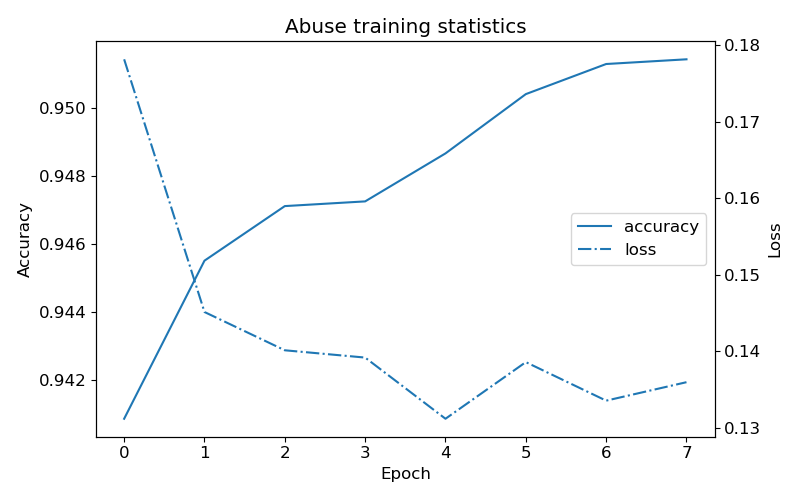}
    \caption{Validation accuracy and loss after each epoch}
    \label{fig:abuse_training}
\end{figure}

Figure~\ref{fig:abuse_training} shows the convergence of
the deep learner for abusive language.

Tables~\ref{tb:product_examples}, \ref{tb:product_examples_manifesto}
and \ref{tb:product_examples_ironmarch} show the highest ranking segments
sorted by predicted intent score, with the corresponding abuse and
abusive-intent scores shown as well. Unsurprisingly in these contexts
there is considerable correlation between intent and abuse.

\begin{table}
  \centering
  \caption{Examples of segments with high abusive intent: Stormfront}
  \label{tb:product_examples}

  \begin{tabularx}{0.49\textwidth}{|l|l|l|X|}
    \hline
    \textbf{Abuse} & \textbf{Intent} & \textbf{Product} & \textbf{Segment} \\ \hline
    0.988 & 0.999 & 0.988 & obama isn t a leftist you ing nazi pig incestuous ing clown i ll rip your ing intestines out and feed them to dogs \\ \hline
    0.963 & 0.996 & 0.959 & first of all i want to address the fact that you are an idiot \\ \hline
    0.931 & 1.000 & 0.931 & i ll ignore the troll are you bnp or anal  \\ \hline
    0.959 & 0.970 & 0.930 & we need segregation from these stupid filthy diseased savages \\ \hline
    0.904 & 0.999 & 0.903 &  don t refer to us as a bunch of hillbillies or we ll kick your ass \\ \hline
    0.901 & 0.999 & 0.901 & we need to stop being soft hiding behind a wall of tolerance and start kicking some black and muslim ass \\ \hline
    0.901 & 0.999 & 0.900 &  if you come to me and threaten my life i will kill you \\ \hline
    0.904 & 0.995 & 0.899 & those white idiots are begging her not to kill black babies i want to buy her a beer honestly if that represents christianity then i want no part of it \\ \hline
    0.898 & 0.999 & 0.897 &  if you tell us to pray quieter we ll kill you \\ \hline
    0.898 & 0.999 & 0.897 &  we ll rape your wife pretoria give me your gun or i ll rape your wife \\ \hline
    0.901 & 0.995 & 0.897 &  about seven racists were outside attempting to kick in the door shouting ing paki im going to kill you black bastard \\ \hline
    0.896 & 0.998 & 0.894 &  i know for a fact that if i had a kid and he was wearing those kind of clothing my reaction would be this one i ll kick your butt you little bastard \\ \hline
    0.904 & 0.988 & 0.893 & i would recommend to say hey don t act a negro who are they for you \\ \hline
    0.897 & 0.993 & 0.890 &  but if you don t i will look for you i will find you and i will kill you \\ \hline
    0.897 & 0.993 & 0.890 & one night when i disagreed with him he d grab me by the throat and said if you don t do what i say i will kill you \\ \hline
  \end{tabularx}
\end{table}

\begin{figure*}[h]
    \centering
    \begin{subfigure}[t]{0.5\textwidth}
      \centering
      \includegraphics[width=\textwidth]{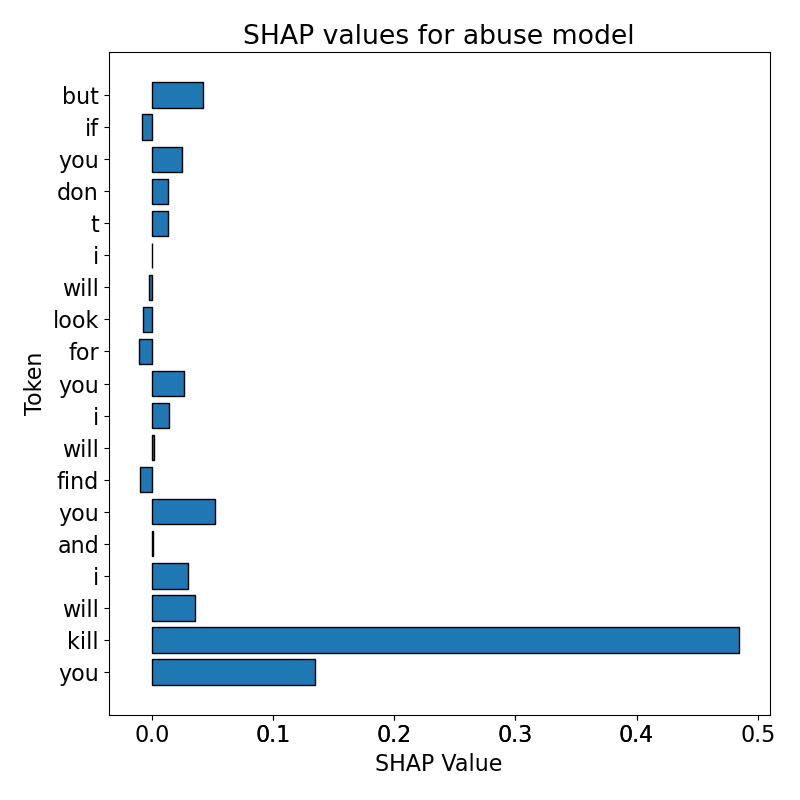}
      \caption{SHAP values for abuse model}
    \end{subfigure}%
    ~
    \begin{subfigure}[t]{0.5\textwidth}
      \centering
      \includegraphics[width=\textwidth]{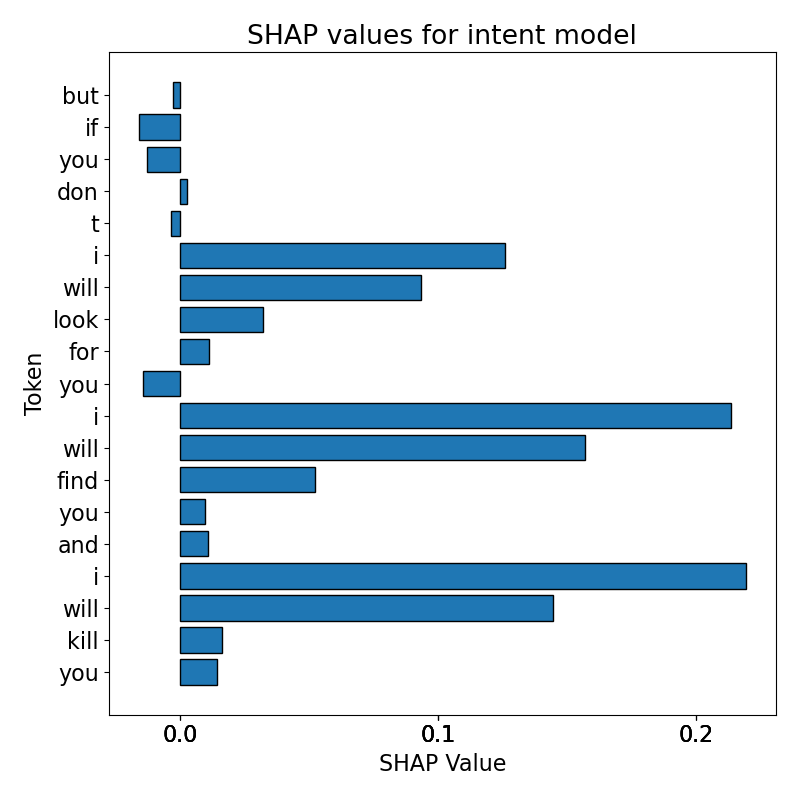}
      \caption{SHAP values for intent model}
    \end{subfigure}

    \caption{Example context with SHAP values for abuse and intent models}
    \label{fig:shap_abuse_intent_2}
\end{figure*}

Figure~\ref{fig:shap_abuse_intent_2} shows Shapley values for the
individual words in an example, illustrating which words have an
impact on the predictions. The abusive language model identifies
``kill you" while the intent model identifies the multiple
occurrences of``I will".

\begin{table}
 \caption{Examples of a range of segments: Manifesto}
 \label{tb:product_examples_manifesto}

  \begin{tabularx}{0.49\textwidth}{|l|l|l|X|}
    \hline

    \textbf{Abuse} & \textbf{Intent} & \textbf{Product} & \textbf{Segment} \\ \hline
    0.968 & 0.999 & 0.967 & we will kill you and drive you roaches from our lands \\ \hline
    0.922 & 0.996 & 0.918 &  if you are released we will find you and kill you if you are in prison we will reach you there if you try to hide these rapist scum we will kill you as well \\ \hline
    0.894 & 0.998 & 0.892 &  i will wipe you the fuck out with precision the likes of which has never been seen before on this earth mark my fucking words \\ \hline
    0.857 & 0.975 & 0.836 &  not only am i extensively trained in unarmed combat but i have access to the entire arsenal of the united states marine corps and i will use it to its full extent to wipe your miserable ass off the face of the continent you little shit \\ \hline
    0.751 & 1.000 & 0.751 & we must crush immigration and deport those invaders already living on our soil \\ \hline
    0.586 & 0.967 & 0.566 &  i will shit fury all over you and you will drown in it \\ \hline
    0.314 & 0.993 & 0.312 &  in the end we must return to replacement fertility levels or it will kill us \\ \hline
    0.285 & 0.980 & 0.280 &  i want your neck under my boot \\ \hline
    0.272 & 0.994 & 0.270 & thus before we deal with the fertility rates we must deal with both the invaders within our lands and the invaders that seek to enter our lands \\ \hline
    0.207 & 0.978 & 0.203 & then i will commit suicide happy in the knowledge i did my best to prevent the death of my race \\ \hline
    0.347 & 0.412 & 0.143 &  these i hate \\ \hline
    0.684 & 0.195 & 0.133 &  would you rather do the killing or leave it to your children \\ \hline
    0.253 & 0.372 & 0.094 & both illegal and legal drug dealers are our racial enemies ruining the health wealth family structure culture and future of our people \\ \hline
    0.347 & 0.228 & 0.079 &  i am a racist \\ \hline
    0.076 & 0.999 & 0.076 &  we must thrive we must march ever forward to our place among the stars and we will reach the destiny our people deserve \\ \hline
 \end{tabularx}
\end{table}

\begin{table}
  \caption{Examples of segments with high abusive intent: Iron March}
  \label{tb:product_examples_ironmarch}

  \begin{tabularx}{0.49\textwidth}{|l|l|l|X|}
    \hline
    \textbf{Abuse} & \textbf{Intent} & \textbf{Product} & \textbf{Segment} \\ \hline
  0.965 & 0.993 & 0.958 & a i m not going to clean that shit up and merge your posts every time you fucking do it if you persist i ll suspend you for a week go reintroduce yourself you faggot joined less than posts who the fuck even remembers you at this point  \\ \hline
    0.953 & 0.998 & 0.951 &  he is a fucking retard and i want to organize a massive troll on him  \\ \hline
    0.910 & 0.999 & 0.909 &  if you suicide i ll kill you  \\ \hline
    0.918 & 0.977 & 0.897 &  once i m done here i m going to spam the fuck out of the forums as a final fuck you  \\ \hline
    0.899 & 0.996 & 0.895 & a i ll end that here because being british sounds snarky as fuck \\ \hline
    0.945 & 0.945 & 0.894 & a anyway i d love to get the fuck away from this cesspit \\ \hline
    0.895 & 0.991 & 0.887 &  if they don t i ll fuck them off \\ \hline
    0.891 & 0.994 & 0.885 &  hahah you re a retard like me then we ll make good company \\ \hline
    0.882 & 0.999 & 0.881 &  i ll leave you to wonder which i am talking about while i blush like an idiot  \\ \hline
    0.906 & 0.960 & 0.870 & a also they pretty much know i ll fuck off to the motherland aftera  \\ \hline
    0.862 & 0.997 & 0.860 &  i like to make it clear i m not a fucking dick sucker that s all but that s how young guys are they idolise anyone who is what they don t dare be \\ \hline
    0.858 & 0.996 & 0.854 &  what the literal fuck sometimes i just want to delete tumblr  \\ \hline
    0.851 & 0.997 & 0.848 &  clean cut but i ll still throw down with niggers \\ \hline
    0.873 & 0.967 & 0.844 &  maybe i will you lazy faggot \\ \hline
    0.841 & 0.996 & 0.838 &  a carrying on with being a british cuck i want to leave the eu so i don t see mongol fuckers like you stealing my money \\ \hline 
  \end{tabularx}
\end{table}

Figure~\ref{tb:windowed_examples} shows the segments making up
entire documents.

\begin{table}
  \centering
  \caption{Examples of entire documents with abuse and intent scores for
each segment. Each group of segments is from one document.}
  \label{tb:windowed_examples}

  \begin{tabularx}{0.49\textwidth}{|l|l|l|X|}
    \hline
    \textbf{Abuse} & \textbf{Intent} & \textbf{Product} & \textbf{Segment} \\ \hline
    0.901 & 0.987 & 0.890 & we need to stop being soft hiding behind a wall of tolerance and start kicking some black and muslim ass \\ \hline
    0.028 & 0.004 & 0.000 & make them uncomfortable in europe by being more open with our nationalistic pride \\ \hline
    \hline
    0.905 & 0.015 & 0.014 & i cant even make it past one or two of these idiots emma west has more guts than all these idiots \\ \hline
    0.028 & 0.981 & 0.028 & to think that she is sitting in jail right now for speaking the truth makes me really mad i want to find her and give her a hug and thank her for being awesome \\ \hline
    \hline
    0.947 & 0.002 & 0.002 & you are all totally stupid \\ \hline
    0.000 & 0.959 & 0.000 & i was submitting a satirical post analogous to all of those which i continue to read on this website \\ \hline
    0.234 & 0.007 & 0.002 & non sensical pretentious etc etc \\ \hline
    0.022 & 0.000 & 0.000 & erm read the title of the thread a typical wn thread \\ \hline
    0.018 & 0.002 & 0.000 & yes you are all terribly bright peoplerolleyes \\ \hline
    0.005 & 0.000 & 0.000 & the post was called posticus b s icus \\ \hline
    \hline
    0.923 & 0.001 & 0.001 & sean taylor is a stupid ape \\ \hline
    0.001 & 0.960 & 0.001 & that i will remember him that way \\ \hline
  \end{tabularx}
\end{table}

\subsection{Comparison with human judgements}

Since the labelling process is inductive, we use agreement with
human assessments as a validation technique.
A web site was created in which volunteers could label segments
as intentful or not, and abusive or not.
Each volunteer received tranches of 5 segments, plus
a qualifying example (not from the dataset and constructed
with a known label).
Tranches in which a user answered incorrectly to the qualifying
question were discarded.
No volunteer received more than 30 examples (6 tranches) to label.

The samples shown to volunteers were sampled from the
extremes of the labelling, that is segments whose labels were
either in [0, 0.4] or [0.6, 1].
A total of 5000 segments were randomly selected and randomly
presented to participants.
Each segment was scored on a first to 3 basis
(that is, 3 consistent votes up to a maximum of 5 total votes) and
removed from the candidate set when it had received 3 consistent
votes.
User predictions were scored as binary (majority intentful or
non-intentful) and weighted (the ratio of votes for intentful
to total votes).

The agreement with the computational predictive model was 80\%
on a binary basis, and 81\% on a weighted basis (that is,
comparing weighted human labelling to real-valued intent
labels).
This suggests that some examples that the intent predictor
found difficult were also ones on which human raters disagreed.

Inter-rater agreement above 70\% are normally considered
adequate, so these results support that the intent predictor
is performing well, especially as intent is a
subjective category for humans.

\begin{figure}[h]
    \centering
    \includegraphics[width=.45\textwidth]{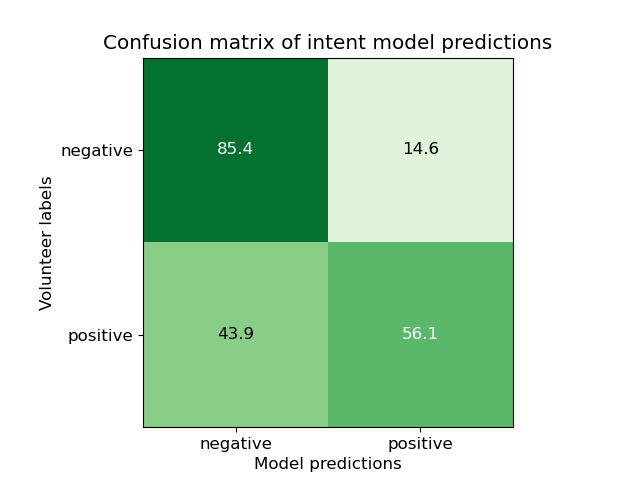}
    \caption{Confusion matrix for intent model using validation labels}
    \label{fig:validation_confusion}
\end{figure}

The confusion matrix between model and human
predictions is shown in Figure~\ref{fig:validation_confusion}.
Raters found considerably more intent than the predictive model
did.
Tables~\ref{tb:false_negative_intent} and
\ref{tb:false_positive_intent} show examples where
the model and human raters disagreed.
In some cases the human raters considered segments
with past tenses as conveying intent (``I decided to join",
``I wanted to feel").
In other cases, they saw what is arguably a weak
form of intent (``I may well respond")
which the model does not see.
Phrases such as ``it going to be splat time" do not contain any
linguistic signals of intent but were plausibly
considered to be (at least potential) intent.

\begin{table}
  \caption{Examples where human raters found intent but the model didn't}
  \label{tb:false_negative_intent}

  \begin{tabularx}{0.49\textwidth}{|l|l|X| }
    \hline
    \textbf{Model score} & \textbf{Human score} & \textbf{Segment} \\ \hline

    0.003 & 1.00 &  well i decided to join the group to educate the morons and counter their ridiculous posts with facts \\ \hline
    0.005 & 0.75 &  old barry might want to think about keeping his ass at home \\ \hline
    0.496 & 1.00 &  the street but its the closest thing we can get to without going through all the legalilties of kicking someones ass in the street \\ \hline
    0.015 & 1.00 &  those words just make me want to break things on you people like you make me sick \\ \hline
    0.427 & 0.75 & i can kill my own spiders and burn my own bodies \\ \hline
    0.001 & 0.75 &  gusts of popular feeling predators and sex objects media portrayals of foreign male and female teachers i wanted to feel her big breasts mr \\ \hline
    0.289 & 1.00 &  quote when you can produce a valid reason for polish immigration into my country rather than your pathetic attempt to justifing it by repeated abuse of other posters then i may well respond to your future posts in the meantime i shall just ignore you \\ \hline
    0.003 & 1.00 &  if one of those filthy rats lays a hand on me it going to be splat time \\ \hline
    0.001 & 0.75 &  make them get off their asses like everyone else has to and work for it \\ \hline
    0.001 & 0.75 &  quote originally posted by maiden america here is what all the gay love and tolerance is leading us to \\ \hline
    0.001 & 1.00 &  all the straight fine young men will have to be exposed to and will be required to put up with the fags \\ \hline
    0.002 & 1.00 &  you and your big stories don t make me come over there and slap you \\ \hline
    0.002 & 1.00 &  now they need to track down and kill those scum \\ \hline
    0.011 & 1.00 &  let your seed grow keep our race alive love your white kin keep them close they will try to sepperate us to destroy us fight for what your beleave stand up for your race and let the jew know we will be on top forever hail victory  \\ \hline
    0.002 & 0.60 & pay attention to what the adl lady says basicly if you dare say anything about a jew your anti smite or racist \\ \hline
  \end{tabularx}
\end{table}

For the examples where the model detected intent but the human scorers
didn't, it is clear that humans do not consider ``we need" or
``we want" to be expressing intent.
However, ``we continue to kick" and ``we are going to be labelled"
are clearly mistaken intent prediction by the model.
These divergences suggest that the model could be improved by
paying more attention to first-person plural pronouns, since
the human scorers clearly believe that these are weaker signals
of intent: ``I need to" is regarded as much stronger than
``we need to".

\begin{table}
  \caption{Examples where the model found intent but the human raters didn't}
  \label{tb:false_positive_intent}

  \begin{tabularx}{0.49\textwidth}{|l|l|X| }
    \hline
    \textbf{Model score} & \textbf{Human score} & \textbf{Segment} \\ \hline
    0.925 & 0.25 &  ie in the case of rhodesia harold wilson the commie bastard linden b johnson the kaffir bottie and verwoed the back stabbing traitor in sa margarate thatcher the slut and the ja stemmers you cowardly kaffir screwing bastards we as member of this forum cannot allow our country s leaders or our nationality divide us we need to unite and watch each other backs for we or the last hope for the homo sapian race \\ \hline
    0.949 & 0.00 & we need to return to medieval era capital punishments for these violent savage animal muds \\ \hline
    0.934 & 0.00 &  we just want to be left alone by races that history has shown can t wipe their own arse without instructions and a tonne of aid money \\ \hline
    0.864 & 0.00 &  we want a world where we can have white heritage month just like the blacks have thier month \\ \hline
    0.805 & 0.40 &  i want the moth r f cker arrested \\ \hline
    0.903 & 0.25 &  my ancestors kick ass and we continue to kick ass \\ \hline
    0.817 & 0.00 &  we are going to be labeled hatred eaten bigots no matter what we do because there are large well funded jew organizations that watch every move we make \\ \hline
    0.946 & 0.40 &  it s seriously starting to grind my nerves and i truly want to know who or what the f is being retarded and bothering me \\ \hline
    0.912 & 0.00 &  jew watch jewish atrocities slave trade jewish slave ship owners quote for decades the white people of america have been subjected to a continual barrage from blacks and others that you and i are somehow responsible for the african slave trade and that we need to atone for our guilt \\ \hline
    0.911 & 0.00 &  we need to be in the middle east like we need a second asshole \\ \hline
    0.605 & 0.00 &  we can speculate all we want except for the cases of obese self esteem issues women but i m convinced there s something important here we re missing along the lines of why well to do pretty white girls lived abusive lives with charles manson and why hetero women with sexual trauma abuse in their pasts become strippers and porn starts and turn bisexual or lesbian \\ \hline
    0.892 & 0.00 & we need that here but the problem is getting people off their asses \\ \hline
    0.891 & 0.00 &  we need a virus that attacks jews and non whites \\ \hline
    0.931 & 0.40 & com but i want to make sure i dont sound like a dumbass \\ \hline
    0.933 & 0.25 &  more like future rapist car thief murderer rapist murderer either way prison or welfare we will support them \\ \hline
  \end{tabularx}
\end{table}

\section{Conclusions}

We address the problem of detecting posts that convey both abusive
content and the intent to act, and so are targets for intelligence
analysts.
We build a predictive system that infers signals of intent from
unlabelled data and a small seed set, using both an n-gram and a
deep-learning approach, acting as colearners. The merged labels
become stable within only a few rounds.
We combine this with a deep-learning biLSTM for abusive language,
and design ways to score individual segments on intent, abuse,
and abusive intent; and score entire documents based on
the scores of their segments.
We appeal to face validation by showing the highest-scoring segments,
and compare the predictions of the model with human scorers,
achieving an agreement of 80\% for class labels and 81\%
for regression.

The methodology developed here can be applied to any other
linguistic property for which an appropriate seed set and
template can be defined.

\bibliographystyle{plain}
\bibliography{sources}
\end{document}